# Multi-Year Vector Dynamic Time Warping Based Crop Mapping

Mustafa Teke[a,b]* and Yasemin Yardımcı[b]

*[a]Scientific and Technological Research Council of Turkey (TÜBİTAK) Space Technologies Research Institute (UZAY), Ankara, Turkey; [b] Informatics Institute, Middle East Technical University, Ankara, Turkey*

*corresponding author: mustafa.teke@tubitak.gov.tr

# Multi-Year Vector Dynamic Time Warping Based Crop Mapping

Recent automated crop mapping via supervised learning-based methods have demonstrated unprecedented improvement over classical techniques. Classification accuracies of these methods degrade considerably in cross-year mapping. Cross-year crop mapping is more useful as it allows the prediction of the following years' crop maps using previously labelled data. We propose Vector Dynamic Time Warping (VDTW), a novel multi-year classification approach based on warping of angular distances between phenological vectors. The results prove that the proposed VDTW method is robust to temporal and spectral variations compensating for different farming practices, climate and atmospheric effects, and measurement errors between years. We also describe a method for determining the most discriminative time window that allows high classification accuracies with limited data. We carried out tests with Landsat 8 time-series imagery from years 2013 to 2015 for the classification of corn and cotton in the Harran Plain of Southeastern Turkey. Besides, we tested VDTW corn and soybean in Kansas, the US for 2017 and 2018 with the Harmonized Landsat Sentinel data. The VDTW method improved cross-year overall accuracies by 3% with fewer training samples compared to other state-of-the-art approaches, i.e. SAM, DTW, TWDTW, RF, SVM and deep LSTM.

## 1. Introduction

The world population is expected to exceed nine billion in 2050 (United Nations 2015). Providing adequate nutrition for the increasing human population is a significant concern. Advanced agricultural technologies, such as precision agriculture and precision irrigation are rapidly emerging to optimise water, fertilisers, and pesticides, thereby enabling higher crop yield. Accurate crop maps are the first requirements of advanced agriculture applications such as yield forecasting. Early-season crop yield estimates are a crucial factor for food security and monitoring agricultural subventions. Crop maps are also an essential tool for statistical purposes to analyse annual changes in agricultural production. However, there are a variety of field crops with similar phenologies and spectral signatures. Likewise, the same plant may have different growing periods in different regions. These properties of field crops render crop mapping a challenge in classification.

Due to the importance of crop mapping on a global scale, various organisations focus on crop monitoring (Rembold and Maselli 2006). One of the most notable examples of crop mapping systems is CropScape. CropScape enables the United States Department of Agriculture (USDA) to map crops in the US for statistical purposes. Governments ensure food security via allocating agricultural subsidies such as LPIS in EU and Turkey (Jansen et al. 2014).

Field surveys are the most basic method of crop mapping. However, they are expensive and may not cover all fields (Esetlili et al. 2018). Furthermore, crop field surveying is prone to human errors (Şimşek, Fatih Fehmi ;Teke, Mustafa;Altuntaş 2016). An effective multi-year crop mapping methodology is required to monitor the status of crops, verify and monitor subventions, forecast crops, ensure price stability and obtain agricultural statistics. Remote sensing is a critical technology that would allow us mapping of field crops by using aerial and satellite imagery from various sources. Crop mapping methods may use single, multi-temporal and time-series satellite imagery. These algorithms typically require field data collection for each year of interest. On the other hand, cross-year crop mapping enables the use of previous field surveys for the present year, thereby reducing the effort required for the collection of training samples.

We surveyed multi-temporal and time-series crop mapping literature with an emphasis on cross-year crop mapping. Land use/land cover (LULC) is an extensively studied research area (Gómez, White, and Wulder 2016; R. Congalton et al. 2014; García-Mora, Mas, and Hinkley 2012). Moreover, crop mapping is a sub-research area of LULC. Multi-temporal and time-series electro-optical satellite imagery were used in the majority of the studies in crop mapping that we surveyed. Multi-temporal images, which are less frequently acquired than time-series imagery, were also commonly used in crop mapping studies. Özdarıcı-Ok and Akyürek developed a method for segment-based classification of multi-temporal Electro-optic and SAR images in Karacabey, Bursa, Turkey (Ok and Akyurek 2012; Ozdarici-Ok and Akyurek 2014). An object-oriented multi-temporal crop classification method for four Landsat 7 ETM+ images of 2012 in Montana, USA was used with an RF classifier to discriminate cereal, pulse, and other classes (Long et al. 2013). Löw et al. developed a decision fusion of decision tree (DT), RF, SVM, and multilayer perceptron (MLP) classifiers to classify multi-temporal RapidEye imagery comprised of alfalfa, cotton, fruit trees, rice, wheat, and melon (Löw, Conrad, and Michel 2015). A winter wheat mapping in northern China with multi-temporal and multi-sensor data was conducted and found out that the RF classifier produced higher accuracies compared to artificial neural networks (ANN), maximum likelihood (ML), and SVM (Liu et al. 2018). These studies performed both training and testing using the same year data. Even though these studies indicated improved accuracies over single imagery, they did not present a multi-year solution due to the acquisition of sparse imagery.

Time-series data involve acquiring a large number of satellite images with high temporal frequency. Petitjean and Weber used DTW for land cover classification with 46 time-series FORMOSAT-2 images of 2006 (Petitjean and Weber 2014). Tatsumi et al. studied classification (alfalfa, asparagus, avocado, cotton, grape, maize, mango, and tomato) of time-series Landsat 7 ETM+ images with random forest (RF) classifier in Peru (Tatsumi et al. 2015). Zheng et al. used 24 time-series Landsat 5 TM and 7 ETM+ of 2010 for classification of crops in Phoenix, AZ (Zheng et al. 2015). Six single crops and three double crops were classified with SVM. Sixteen-day MODIS time-series data of 2001 was used for land use classification (urban, forest, agriculture) in the USA (Shao and Lunetta 2012). SVM, neural networks (NN) and classification and regression tree (CART) were compared and SVM performed better. Lunetta et al. classified wheat, corn, and cotton with MODIS time-series 16-day NDVI composite data in the Great Lake Basin, US, and Canada with a three-layer MLP classifier. In their study, crop layers of 2005-2007 were compared with USDA NASS' Cropland Data Layer (CDL) and crop ration patterns were analyzed: observed differences between CDL and their results were between 1-11.1%. Zong et al. developed a spectro-temporal crop classification method to classify corn and soybean from time-series Landsat 5 and 7 images (Zhong, Gong, and Biging 2014). Time-series features are extracted from the parameters of a double sigmoid curve. This study achieved 89.4% same-year (SY) accuracy and 83.4% cross-year (CY) accuracies. Maus et al. proposed time-weighted dynamic time warping (TWDTW), which is an improvement over dynamic time warping (DTW) by incorporating the time difference between samples as an additional cost (Maus et al. 2016). In another study, pixel-based and object-based TWDTW methods were compared with random forest (RF) with Sentinel-2 time-series data. Object-based TWDTW achieved comparable results to RF classifier (Belgiu and Csillik 2018). Massey et al. studied the multi-year distribution of major crop types in the conterminous US with time-series MODIS data between 2001 and 2014 (Massey et al. 2017). A phenology-based decision tree approach achieved year-specific (same-year) accuracies of >78% and generalized (multi-year) accuracies > 75% in 13 agro-ecological zones. A multi-decade and multi-sensor time-series crop mapping was performed in (Pringle, Schmidt, and Tindall 2018) with Landsat imagery was fused with Sentinel-2 and MODIS when available. The method composed of two export-rules and two-random forests to classify winter and summer crops in Queensland, Australia between 1987 and 2017.

A majority of the studies on multi-temporal or time-series satellite imagery crop classification did not take time into account as a feature and focused on mapping crops using same-year data for both training and validation. However, multi-year analysis enables earlier classification of crops based on previous years' data. Only a limited number of studies conducted multi-year comparisons such as (Zhong, Gong, and Biging 2014; Maus et al. 2016). These studies presented classification accuracies where cross-year results were considerably lower compared to same-year results, and they required a substantial amount of training samples. Even if cross-year crop mapping eliminated the necessity of yearly training sample collection, these studies still needed considerable training data. Again, most studies did not incorporate annual temporal variations in their studies; one notable exception is the work of (Maus et al. 2016). RF and SVM were most used classifiers in crop mapping.

Furthermore, we considered deep (DL) learning methods for cross-year crop mapping. DL has gained popularity in recent years due to its applications in numerous areas (Lecun, Bengio, and Hinton 2015). Deep convolutional neural networks and recurrent neural nets were applied for crop mapping (Kamilaris and Prenafeta-Boldú 2018; Liakos et al. 2018). DL methods achieved higher classification accuracies compared to other classification methods such as SVM and RF (Kussul et al. 2017). However, DL requires a vast amount of training data and an extensive amount of computing power. The crop-mapping studies that used DL, mentioned above, were tested with only same-year data, while the majority of the data were used in training.

In this study, we aim to develop an efficient cross-year crop mapping algorithm which uses a limited number of training samples and is resistant to annual measurement and growth variations.

The main contribution of the study is the development of a novel vector distance-based optimal time-warping algorithm (VDTW). The VDTW method overcomes difficulties in cross-year crop classification in which training and test data are selected from different years: spectral shifts due to changes in illumination at the time of observation and temporal shifts in growth due to yearly climate variations or farming practices. We simulated different cases of illumination and growth changes. Furthermore, we tested our methodology in a multi-year approach in two regions with distinct cropping practices. The proposed approach requires fewer training samples compared to other methods; thus, it significantly reduces the costly collection of field data.

As a second contribution of this work, we focused on the feasibility of exploiting crop phenologies to use fewer and effective image acquisitions. A method which automatically determines the optimal time window in which crops have discriminative phenological features is developed. The algorithm developed to select this optimal time allows mid-season crop classification, enabling early accurate prediction of crop yields. In this way, the necessary precautions for transport, storage as well as price volatility could be taken.

## *2. Study Sites and Data*
### *2.1. Study Sites*

In this study, we tested the VDTW method in two different regions: The Harran Plain. The Harran Plain is located in the South East of Turkey. The location of the Harran Plain is shown in Figure 1. The region has a Mediterranean climate with about 400-

450mm yearly rainfall according to General Directorate of Meteorology of Turkey.

The Harran Plain is bordered by Şanlıurfa city and Germüş Mountains in the north, Tek Tek Mountains in the east, Akçakale town and the Syria border in the south and Fatik Mountains in the west. Its length is 65 km from north to south, and its total area is 225000 ha. According to the Turkish Statistical Institute (TUİK) data, barley, wheat, corn, and cotton are the principal crops of the Harran Plain. The Harran Plain is irrigated by the canals from the Atatürk Dam (Ozdogan et al. 2006; Çelik and Gülersoy 2013).

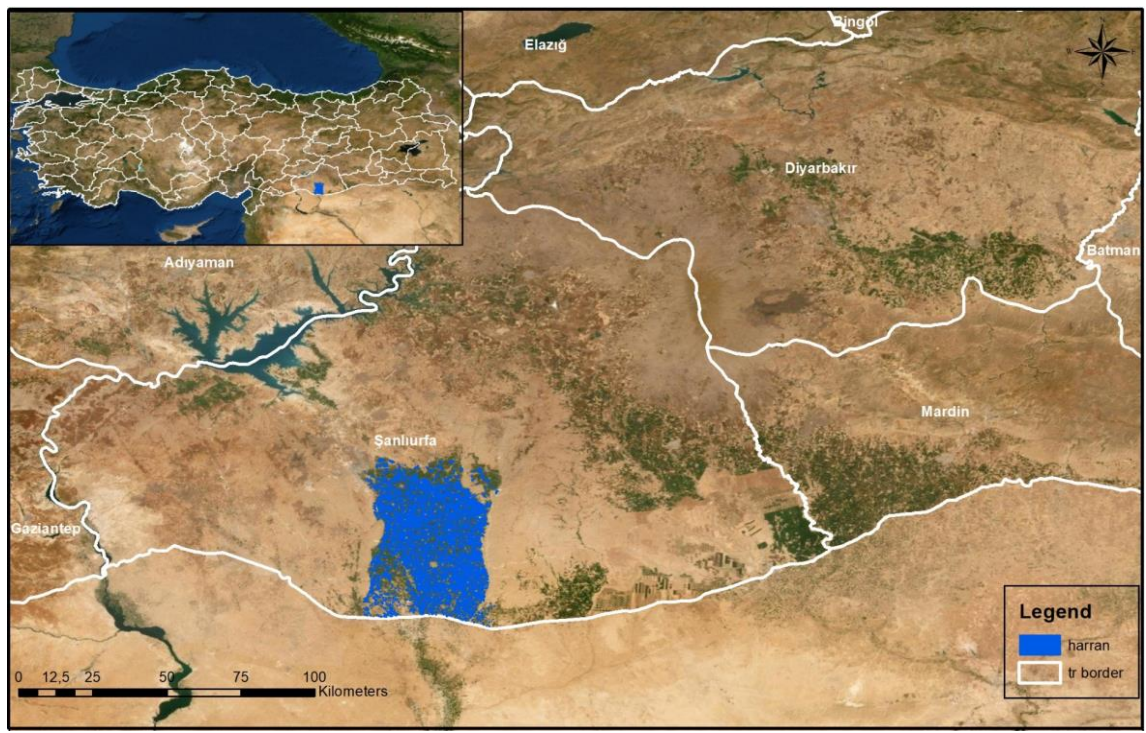

Figure 1. The Harran is depicted in Turkey.

Our 2nd dataset was selected from the Northeast of Kansas (Figure 2). The Kansas data set extends on Brown, Jackson, Nemaha, Shawnee, Pottawatomie, and Wabaunsee counties. Major crops in the region are corn and soybean.

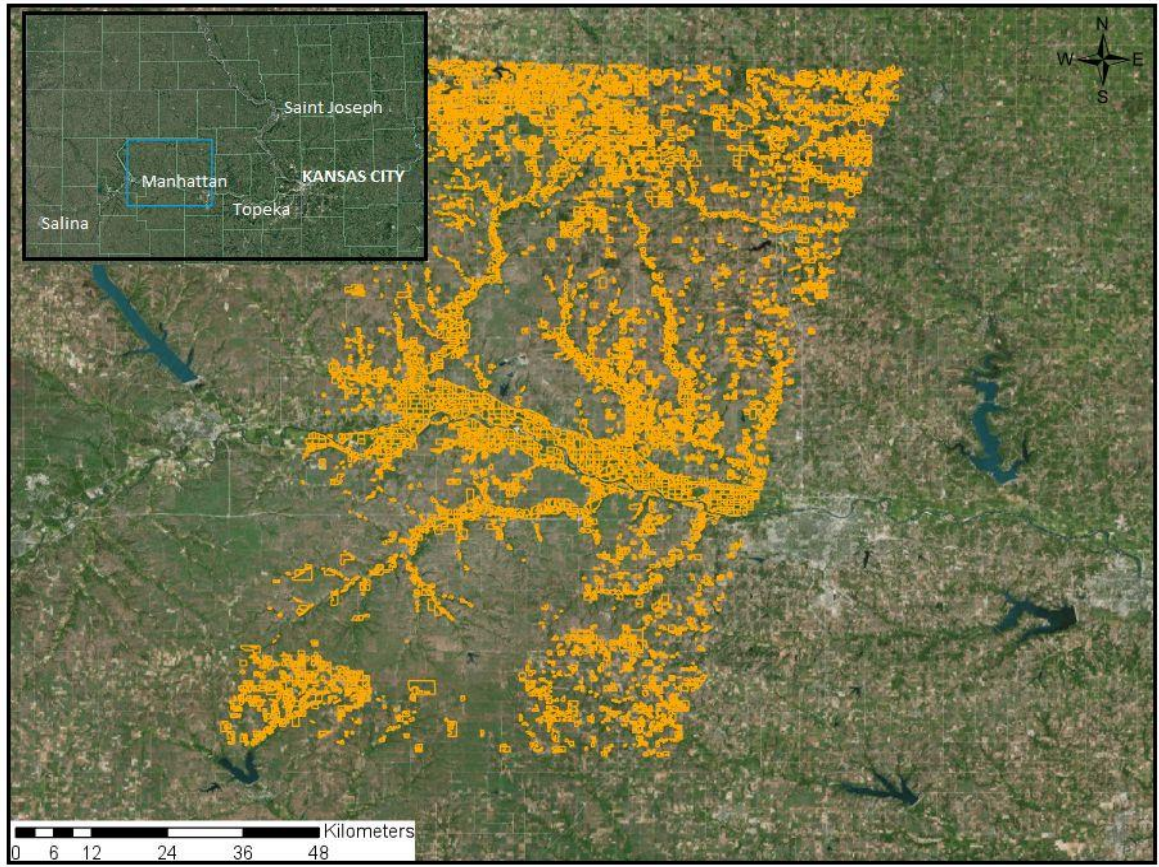

Figure 2. The Kansas dataset is depicted.

## 2.2. Satellite Imagery

Multi-year data of Landsat 8 satellite were used in this study. Landsat 8 covers the Earth every 16 days. Landsat 8 data were converted to surface reflectance by the U.S. Geological Survey (USGS) (Landsat 8 Surface Reflectance Product Guide v1.2 2015). Harmonized Landsat Sentinel data were used for the Kansas dataset.

The imagery of the Harran Plain was from early June to the end of October. Twenty images from 2013 and 2015 and 19 images from 2014 were used. Imagery acquisition details for the Harran Plain are presented in Figure 3.

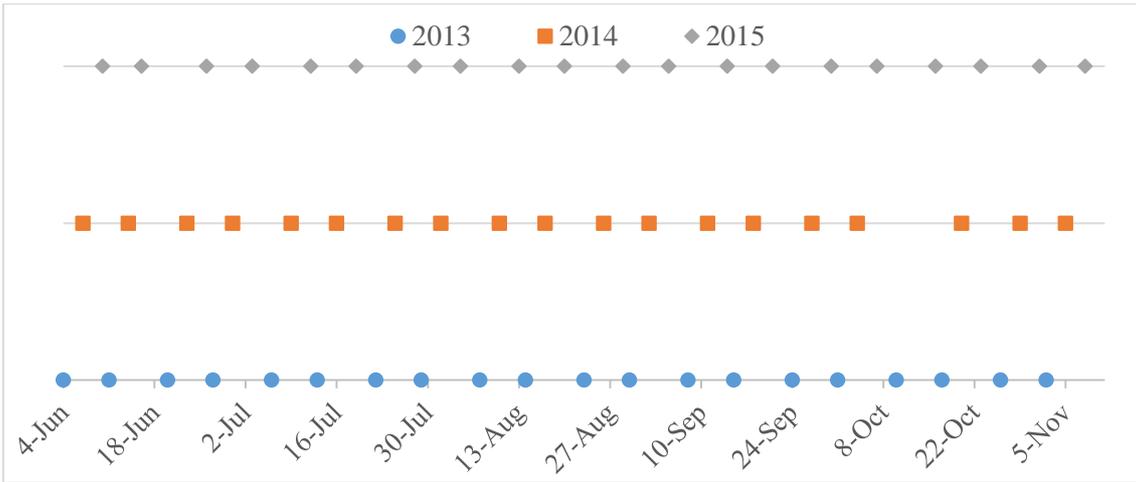

Figure 3. Landsat 8 imagery Harran dataset acquisitions dates in 2013, 2014 and 2015.

We used Harmonized Landsat 8 and Sentinel-2 (HLS) data for the Kansas dataset. The HLS data enabled more cloud-free data acquisitions. The Kansas data set has 22 images ( 15 Landsat 8 and seven Sentinel-2 ) in 2017 and 22 ( five Sentinel-2 and 17 Landsat 8) images in 2018. Harmonized Landsat Sentinel project resamples Sentinel-2 imagery in to match Landsat 8 in spatial and spectral properties(Claverie et al. 2018).

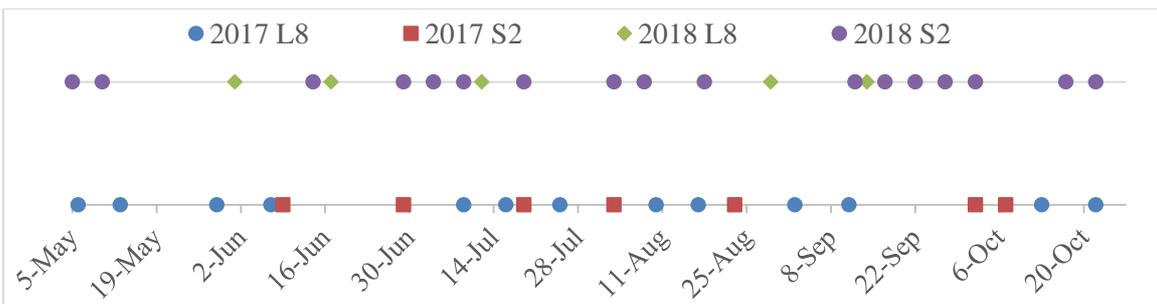

Figure 4. Landsat 8 and Sentinel-2 Kansas dataset imagery acquisition dates in 2017 and 2018.

## 2.3. Ground Truth Data

Ground truth is based on the Ministry of Agriculture and Forestry's National Registry of

Farmers (NRF, Turkish: Çiftçi Kayıt Sistemi, ÇKS) for Turkey. In the NRF, farmers declare the crops that they will grow in order to apply for government agricultural subsidies (Yomralioglu et al. 2009). On the other hand, the ground truth of the Kansas dataset is based on USDA NASS's the Cropland Data Layer (CDL). The CDL data was created based on USDA's Farm Services Agency (FSA) Common Land Unit (CLU) data.

The NRF contains vectors of agricultural fields. Regarding the GT, we started with census data: the declaration from the National Registry of Farmers. In the case of Kansas dataset, we used CLU 2008 data as field boundaries.

The median vegetation index (VI) time-series vector data of each field is assigned as a sample in the tests. A summary of the characteristics the Harran Dataset is presented in Table 1 and the Kansas dataset is depicted in Table 2.

Table 1. Number, percentage distribution and areas of corn and cotton fields in the Harran dataset in 2013, 2014 and 2015

|      |        | #Fields | %Samples | Area (ha) |
|------|--------|---------|----------|-----------|
|      | Corn   | 1192    | 21.8     | 12366     |
|      | Cotton | 4285    | 78.2     | 43968     |
| 2013 | **Total** | 5477 |          | **56333** |
|      | Corn   | 692     | 13.2     | 7321      |
|      | Cotton | 4561    | 86.8     | 47395     |
| 2014 | **Total** | 5253 |          | **54716** |
|      | Corn   | 517     | 15.4     | 5863      |
|      | Cotton | 2849    | 84.6     | 31094     |
| 2015 | **Total** | 3366 |          | **36957** |

Table 2. Number, percentage distribution and areas of corn and soybean fields in the Kansas dataset in 2017 and 2018

|  |  | #Fields | %Samples | Area (ha) |
|---|---|---|---|---|
| 2017 | Corn | 1167 | 41.66 | 67952 |
|  | Cotton | 4083 | 58.34 | 89479 |
|  | **Total** | 5250 |  | **157431** |
| 2018 | Corn | 2307 | 42.99 | 71714 |
|  | Cotton | 3059 | 57.01 | 86913 |
|  | **Total** | 5366 |  | **158627** |

## 3. Methods

### *3.1 Phenological Crop Classification*

Growth and status of the crops are measured by vegetation indices. The proposed VDTW algorithm is based on time-series vegetation phenology. Vegetation indices such as NDVI are used to measure phenology information. The same region may be observed at different observation angles. Thus, phenology information extracted from different viewing angles was affected by the bidirectional reflectance distribution function (BRDF). Huete et al. found that the soil-adjusted vegetation index (SAVI) is resistant to BRDF effects compared to NDVI (Huete et al. 1992). Modified soil-adjusted vegetation index (MSAVI) improves SAVI since it automates the calculation of the soil line. In this study, NDVI, SAVI, optimised SAVI (OSAVI), EVI, enhanced NDVI (ENDVI), and WDRVI were compared. MSAVI slightly improved VDTW method's cross-year crop mapping overall accuracy by 0.05% compared to NDVI (results not shown here). Moreover, the use of MSAVI improved the cross-year overall accuracy of VDTW method in the Kansas dataset by 1.5%(EVI) to 2%(NDVI).

The crop calendars of corn, cotton, and soybean in the Harran Plain and Kansas(of Agriculture-National Agricultural Statistics Service 2010) are presented in Figure 5. Figure 6 shows variations in phenology MSAVI values of corn and cotton have large differences in their early growth periods samples in 2013 as a box plot. The growth is followed by a steady maturity period. Also, harvests of corn and cotton highly overlap: corn is harvested earlier than cotton. Corn is grown after the harvest of winter wheat in the Harran Plain as the second crop (double cropping).

|  | Months | March | April | May | June | July | August | September | October | November |
|---|---|---|---|---|---|---|---|---|---|---|
| Harran | Corn |  |  |  |  | Sowing | Growth |  | Harvesting |  |
| Harran | Cotton |  | Sowing | Growth |  |  |  | Harvesting |  |  |
| Kansas | Corn |  |  | Sowing | Growth |  |  | Harvesting |  |  |
| Kansas | Soybean |  |  |  | Sowing | Growth |  |  | Harvesting |  |

figure 5. Crop calendars of corn and cotton in the Harran Plain and Corn, Cotton; and corn and soybean in Kansas.

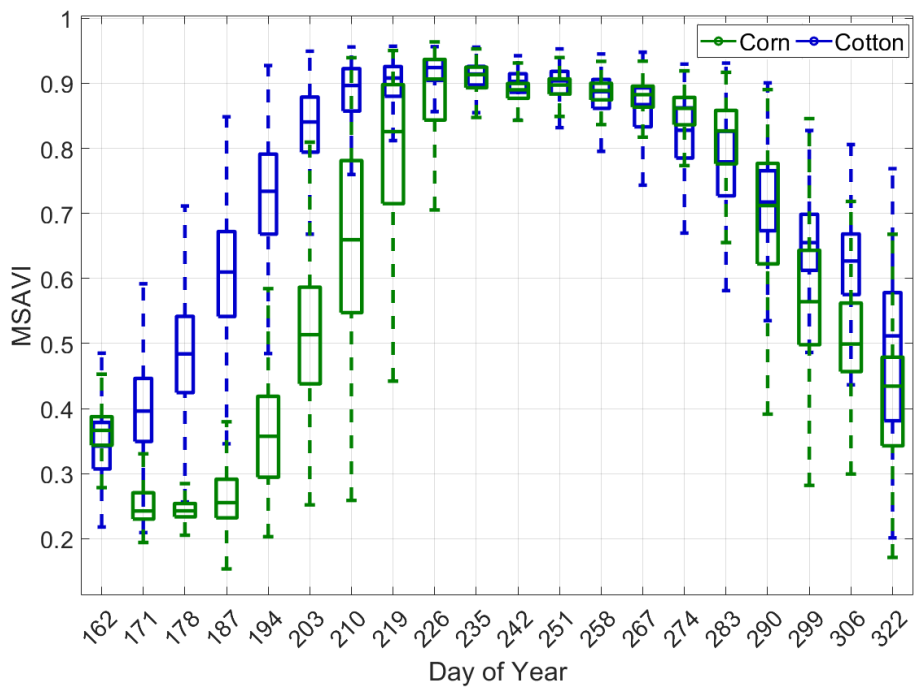

**Figure 6**. Box plots at each image acquisition in 2013 show variations in vegetation phenologies of all corn and cotton samples for the Harran dataset. Central mark

indicates the median values. Bottom and top edges show 25$^{th}$ and 75$^{th}$ percentiles, respectively.

### 3.2     *Multi-year Crop Mapping Approach*

An approach having high cross-year classification accuracy will benefit multi-year crop mapping studies. However, selecting training samples for each distinct location and crop at each year is difficult. In this study, a scheme is developed to classify crops with data from cross-years: training data are selected from one year while tests are performed with another year's data.

The proposed approach aims to present an efficient multi-year classification methodology as we incorporate cloud information cloning and time-series data smoothing.

A summary of the algorithm steps is presented in Figure 7. The vegetation index is computed from radiometrically corrected time-series satellite images. Atmospheric or illumination effects may degrade the performance of times series classification methods. Data smoothing methods have been used to reduce these effects (Arvor et al. 2008), and in our work, we smoothed our (time-series) data by the Savitzky-Golay (SG) filtering method(Kim et al. 2014).

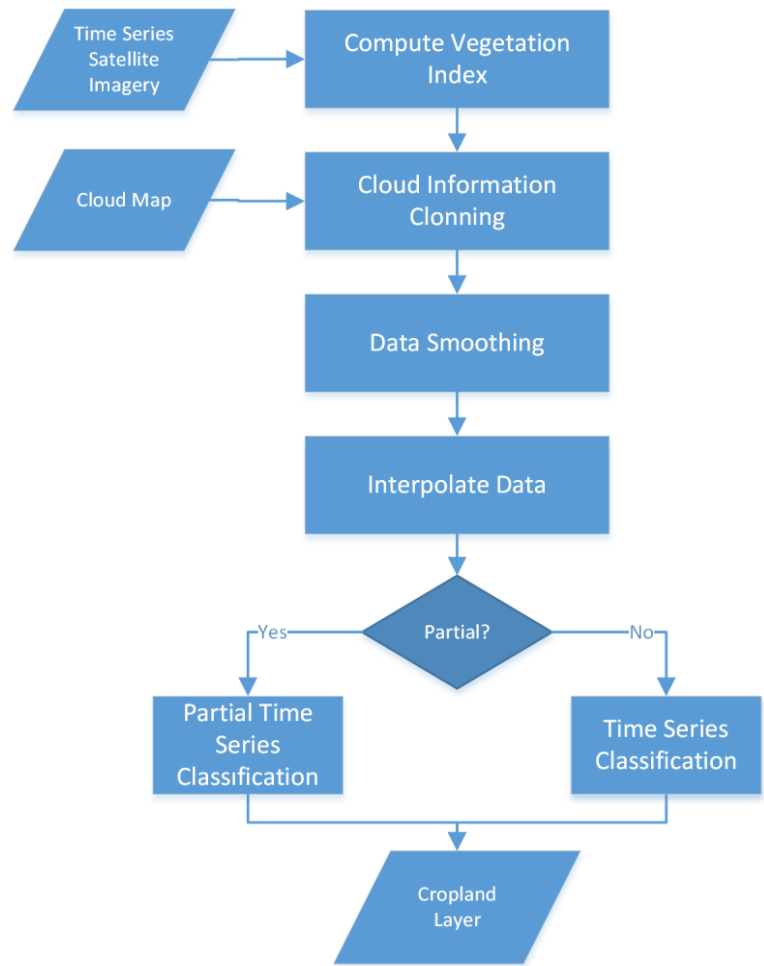

Figure 7. Multi-year Time-series Classification Algorithm Steps

Landsat 8 cloud and shadow masks are produced by the Fmask algorithm (Zhu and Woodcock 2012). We applied cloud information cloning: cloudy samples were linearly interpolated with the Inverse Distance Weighting (IDW) method by using the nearest two cloud-free images (Kalkan and Maktav 2018). Even though the Fmask algorithm could detect clouds successfully, it may not detect cloud shadows as effectively. However, we found the Fmask algorithm and SG smoothing, followed by median of time-series field phenology was adequate for successful classification. Kansas dataset uses the Harmonized Landsat 8 and Sentinel-2 satellite imagery (HLS). However, the Fmask algorithm (Fmask v3) which were used in HLS data is not optimal with Sentinel-2 data. This resulted in missed shadows and clouds in some cloudy

Sentinel-2 scenes. Moreover, we used Fmask version 4, which improved shadow and cloud detection with Sentinel-2 data. We used double sigmoid fitting instead of SG of crop phenologies to remove the remaining artefacts.

To enable cross-year classification, data are linearly interpolated between $[t_l, t_u]$ where $t_l$ and $t_u$ are the lower and upper limits of the time-window. Time-series classification is used to classify with same-year or cross-year classification. Optionally, data is classified with a partial time-series approach. Finally, a cropland layer is produced showing the classification results.

### *3.3    Vector Dynamic Time Warping: VDTW*

We studied phenological variations of a crop in various experimental settings within our datasets. For this purpose, we simulated crop signatures to analyse the behaviour of DTW and spectral angle mapper (SAM) methods. Spectral angle mapper is a commonly used measure in hyperspectral image analysis describing the angular distance between two spectra (Kruse, Lefkoff, and Dietz 1993). Dynamic Time Warping (DTW) is a technique that finds the optimal alignment in translation and scaling between two time-series data sequences (Müller 2007). The sequences are matched in by dynamic programming to find optimal distances between signals. Shift and scale are simulated in different scenarios. These scenarios are given in the Appendix The scenarios aim to simulate farmers' practices, illumination, and climate changes both in the same-year and cross-years.

Our analyses with the simulations showed that both DTW and SAM have disadvantages while dealing with time-series phenological data since phenological measurements of crops at different dates may vary, as the weather and illumination conditions are not static.

We propose a new method, which is both robust to shift in crop growth and illumination differences: Vector Dynamic Time Warping (VDTW). While DTW is based on Euclidean distance $d$, we propose to use angular distance $a$ as shown in Figure 8. VDTW computes the optimal warping path of spectral distances between two phenological observations. The length of the DTW search window should consider the possible phenological variation between years. According to our experiments with various window sizes, the search window size was set to ± 15 days.

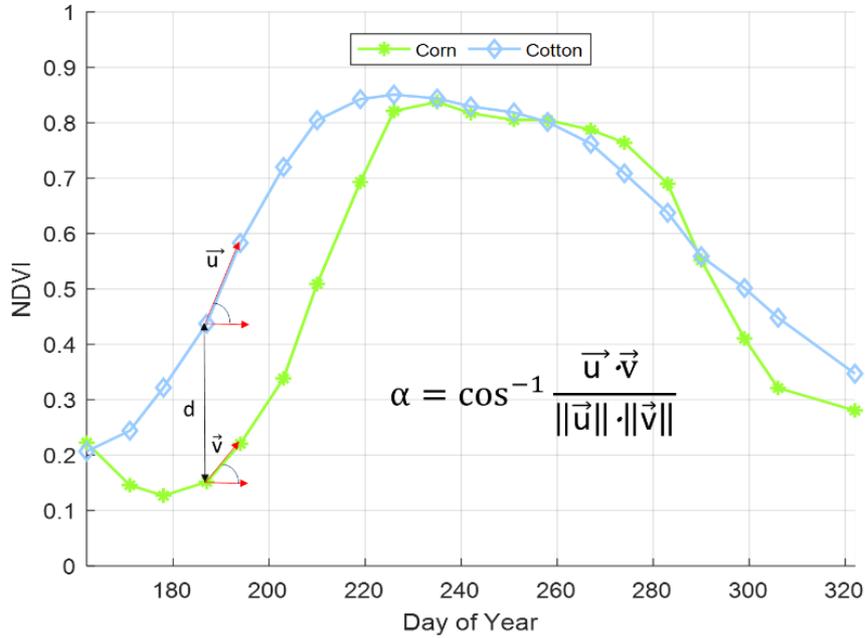

**Figure 8.** Angular distance metric between phenology of two crops at an observation date

The first step in VDTW algorithm is constructing $n$-by-$m$ distance matrix $\Psi$ whose elements $\psi_{i,j}$ is computed as the angle α between $\vec{u}_i \in \cup \; \forall \; i = 2, \ldots, n$ and $\vec{v}_i \in \cup \; \forall \; j = 2, \ldots, m$. $\vec{u}_i$ and $\vec{v}_i$ are unit vectors at each element. $\psi_{i,j}$ is computed as follow:

$$\psi_{i-1,j-1} = cos^{-1} \frac{\vec{u}_i \cdot \vec{v}_j}{\|\vec{u}_i\| \cdot \|\vec{v}_j\|} \tag{1}$$

The accumulated ($m - 1$ x $n - 1$) distance matrix, D, is computed from ψ by calculating the recursive sum of distances:

$$d_{i,j} = \psi_{i,j} + min\{d_{i-1,j-1}, d_{i-1,j}, d_{i,j-1}\} \quad (2)$$

Computation is subject to following boundary conditions:

$$d_{i,j} = \begin{cases} \psi_{i,j} & i = 1, j = 1 \\ \sum_{k=2}^{i} \psi_{k,j} & 2 < i \leq n-1, j = 1 \\ \sum_{k=2}^{j} \psi_{i,k} & i = 1, 2 < j \leq m-1 \end{cases} \quad (3)$$

The source code of the VDTW method is available at GitHub: https://github.com/mustafateke/VDTW

### 3.4 Partial Time-Series Classification

A new method is presented in the previous section. In this study, we also propose a partial time-series approach which achieves high classification accuracies with fewer data acquisitions, using distinct time-periods of crop phenologies.

For example, corn and cotton in the Harran Plain are sown at a specific date; however, they both start to have the same phenological properties beginning from mid-August, after which their growths are nearly the same. Corn and cotton can be discriminated in their early growth until mid-August (Figure 6). Our method exploits this phenologically invariant region for improved cross-year crop classification.

The partial time-series algorithm has three significant steps. The algorithm finds the optimal classification window around the pivot day.

*Algorithm steps:*

First, the pivot day where the difference between the vegetation index (VI) of crops is maximum is determined (Figure 9(a)). Median values of all samples from each crop are used in this computation. The pivot day is determined as:

$$J^* = arg \max_{t_l < J < t_u} abs(VI_{C_1}(J) - VI_{C_2}(J)) \quad (4)$$

where $J^*$ denotes pivot day, $C_1$ and $C_2$ are first and second crops, $t_l$ denotes the minimum common day and, $t_u$ denote the maximum common day shared by time-series data of all years.

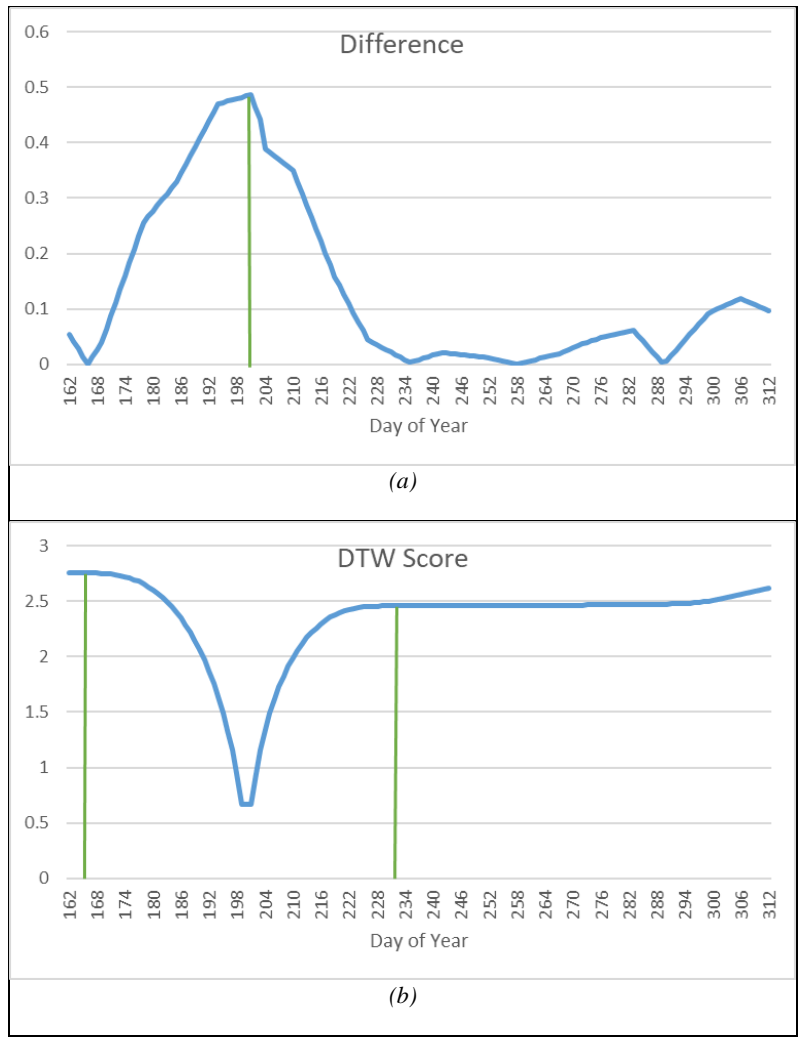

(a)

(b)

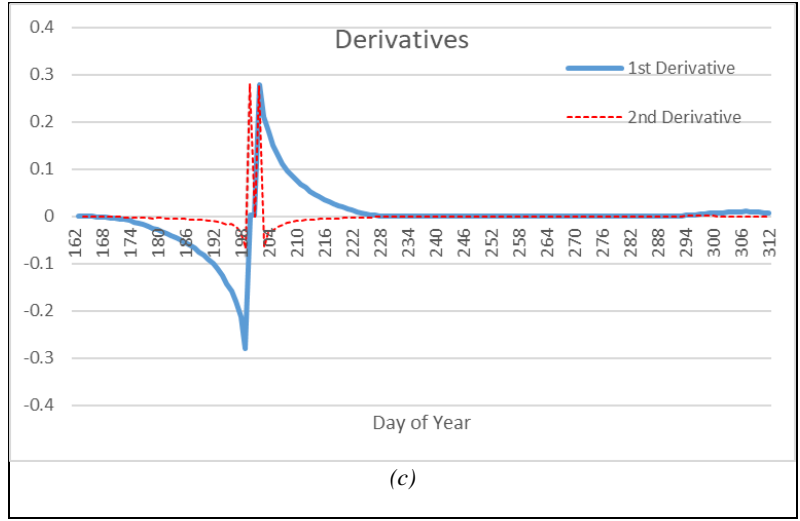

*(c)*

Figure 9. (a) The maximum difference of VI values between corn and cotton, (b) DTW scores between corn and cotton centred on the pivot day expanding on both sides, (c) First and second derivatives of DTW scores.

DTW scores of vectors extending in both directions are computed by centring the pivot day (Figure 9(b)). Lower DTW scores correspond to higher similarity. The increase in DTW scores is steady after specific periods, which coincides with discriminative regions of corn and cotton.

$$Score(J) = DTW(VI_1([J^*,J]), VI_2([J^*,J])) \tag{5}$$

where $t_l < J < t_u$.

We find the first days from the pivot by extending to initial and final dates until first and second derivatives are zero (Figure 9(c)). First and second derivatives indicate that DTW scores are steady after these days, as a result determining the boundaries of the optimal time window.

$$find\ score(J)' = 0\ AND\ score(J)'' = 0 \tag{6}$$

The optimal time window [o1, o2] for classification of corn and cotton are computed as day 170 and day 227, corresponding to mid-June and mid-August. In the case of three or more crops, each crop is compared to others and the minimum length time window is selected against other crops.

## 4. Results

In our study, we used the median of each field as a sample. A stratified random selection strategy was applied to training sample selection (Olofsson et al. 2014), and selected training samples were excluded from the test samples in same-year tests. The same training samples for each test are used in training for all methods. Tests were repeated 100 times to minimise the effect of non-representative outlier samples such as crops grown too early or too late. We compared different methods against various numbers of training samples to evaluate their performance with a limited number of training samples. Congalton suggested using at least 50 samples from each class when the number of classes is less than 12 (R. G. Congalton 1988). We varied the number of training samples in 5,10,…,50 based on their findings.

Detailed tests are performed for same-year and cross-year classification accuracies. Double sigmoid features with RF classifier and SVM, Time-series (VDTW, SAM, and DTW, TWDTW) and partial time-series (PVDTW) were compared in this study. RF classifier contains 1000 trees. SVM has the RBF kernel, and its parameters are selected after an extensive grid search of cross-validation of training samples. As DL methods gained much attention in classification, we used two-layer deep long short-term memory (LSTM) with 100 units at each layer followed by a softmax layer (Reimers and Gurevych 2017).

In our cross-year tests, the Harran dataset is the most challenging since corn and cotton's phenologies vary each year after peak growth until the harvest. Same-year and cross-year percent overall accuracy scores of tested methods are shown in Table 3. Our tests have shown that VDTW provides the highest overall accuracies both in the same year at 99.22% (Figure 10) and cross-year at 98.29% (Figure 11).

SAM and RF methods had similar accuracies in the same year; however, RF was not robust to growth changes in the cross-year as SAM. SAM cross-year scores were below 94.78%. RF was able to reach 94.45% with a maximum number of training samples. VDTW was more robust to shifts in growth and changes in illumination compared to other methods. The best two performers VDTW and TWDTW achieved the cross-year 50 training sample and 100-replication overall average classification accuracies of 98.29% and 95.29%, respectively. The 95% confidence interval for overall accuracy differences between VDTW and TWDTW methods were between 2.77% and 3.23%. (Table 3). TWDTW's time cost improved DTW's cross-year overall accuracy from 93.31% to 95.29%. The effect of window size of VDTW and DTW was investigated by extensive runs. Even though window size makes a difference in the accuracy, VDTW was always superior. Time series with Deep LSTM initially produced lower accuracies for training sample size < 20 for each class. Deep LSTM obtained similar overall accuracies with DTW and SAM for the training sample size of 50(~%1 of samples) for each class.

Tests with a varying number of training samples revealed that VDTW maintained high classification accuracies with a fewer number of samples compared to other methods as shown in Figure 10. In other words, the advantage of the proposed approach is its ability to attain high classification accuracy independent of the training set size.

Partial time-series applied to VDTW also achieved similar accuracy values as the core method. Partial time-series the applied version of VDTW, PVDTW, reduces the amount of data by using fewer data limited by time windows. These time windows are based on phenological differences between crops.

RF and SVM classifiers, which use features extracted from time-series data, have lower performance than other methods in the tests. Performances of RF and SVM are lower since curve fitting is designed for single cropping and may not always fit the optimal curve for double cropping case. Time-series methods such as proposed VDTW, SAM, and DTW are robust to double cropping cases.

Finally, we tested the VDTW method with data from Kansas. Crops in Kansas are distinctly grown. Same year crop mapping accuracies were high for all classifiers, as shown in Table 4. TWDTW method obtained highest the same-year overall accuracy of 99.02% followed by VDTW and LSTM having overall accuracies of 98.74% and 98.60%. On the other hand, VDTW resulted in higher overall accuracies than TWDTW by 1.72% and other methods in the cross-year tests.

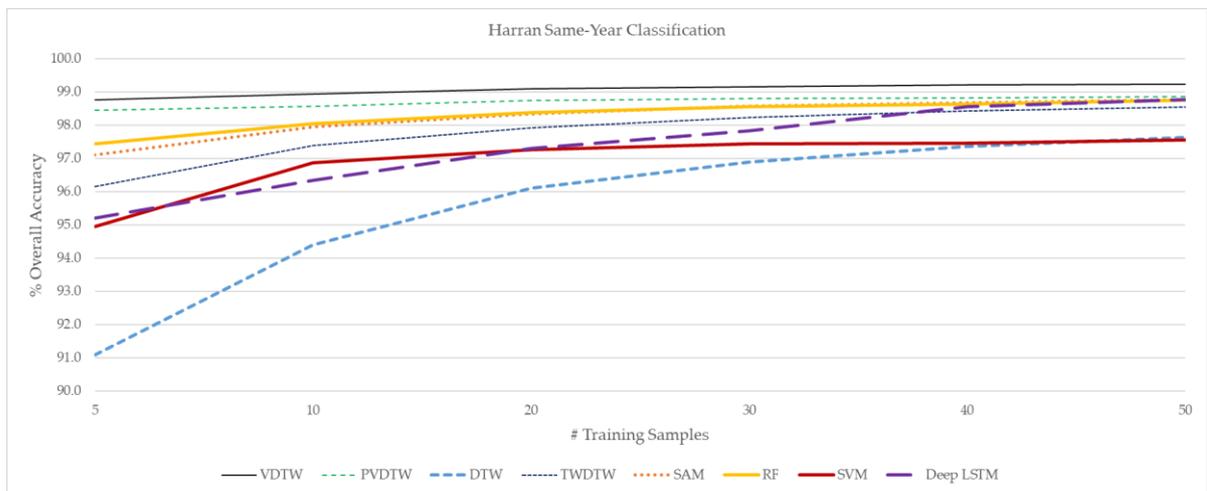

Figure 10. Harran dataset same-year classification results at various training sample sizes.

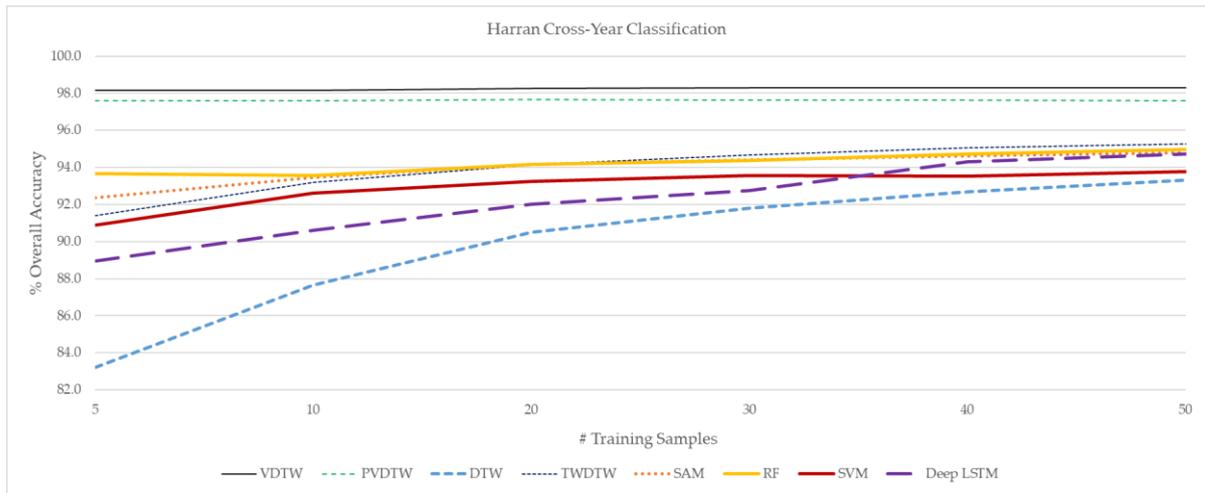

Figure 11. Harran dataset same-year classification results at different training sample sizes.

Table 3. Percent average overall accuracies of proposed and compared methods with 50 samples from each class for the Harran Dataset. Samples are selected with the stratified random selection

|  | VDTW | PVDTW | DTW | TWDTW | SAM | RF | SVM | Deep LSTM |
|---|---|---|---|---|---|---|---|---|
| Same-year | **99.22** | 98.86 | 97.64 | 98.54 | 98.77 | 98.72 | 98.36 | 98.76 |
| Cross-year | **98.29** | 97.58 | 93.31 | 95.29 | 94.78 | 94.45 | 92.40 | 94.74 |

Table 4. Percent average overall accuracies of proposed and compared methods with 50 samples from each class for the Kansas Dataset. Samples are selected with the stratified random selection

|  | VDTW | PVDTW | DTW | TWDTW | SAM | RF | SVM | Deep LSTM |
|---|---|---|---|---|---|---|---|---|
| Same-year | 98.38 | 97.57 | 78.53 | 98.63 | 98.42 | 98.35 | 97.91 | 98.30 |
| Cross-year | 89.68 | 84.89 | 73.01 | 88.40 | 85.55 | 86.06 | 86.26 | 87.10 |

User's accuracy and producer's accuracy for the Harran dataset are presented in Table 5. User's accuracies are similar for both crops; however, several mislabelled corns result in lower producer's accuracy for corn. Both user's and producer's accuracies of cotton are over 99% in same-year tests and 98% in cross-year tests.

Table 5. Average User's Accuracy and Producer's Accuracy of VDTW classification results with 50 samples for the same-year and cross-year.

|  | User's Accuracy | | Producer's Accuracy | |
|---|---|---|---|---|
|  | **Corn** | **Cotton** | **Corn** | **Cotton** |
| **Same-year** | 97.29 | 99.57 | 97.90 | 99.48 |
| **Cross-year** | 95.28 | 99.16 | 95.73 | 98.72 |

High user's accuracy both in the same and cross-year tests show that misclassification percentage of corn and cotton is low. However, low user's accuracy of corn indicates that 4.72% of corn is labelled as cotton in cross-year tests. Misclassification error is 2.71% in the same-year tests. Kappa values were 0.97 for the same-year tests and 0.94 for the cross-year tests.

A detailed view of VDTW classification is provided in the form of confusion tables. Confusion matrix in Table 6 shows the number of fields which were correctly classified as corn and cotton with training data from the same or other years. Cotton was correctly classified while some percent of corn is misclassified as cotton.

Table 6. Average Confusion Matrix of 100 tests for VDTW Classification with 50 samples in the Harran Plain. Columns are observations while rows are predictions. Years in rows are training and years in rows are test years.

|  |  | 2013 | | 2014 | | 2015 | |
|---|---|---|---|---|---|---|---|
|  |  | **Corn** | **Cotton** | **Corn** | **Cotton** | **Corn** | **Cotton** |
| **2013** | **Corn** | 1108 | 34 | 619 | 13 | 513 | 4 |
|  | **Cotton** | 34 | 4201 | 73 | 4548 | 4 | 2845 |
| **2014** | **Corn** | 1158 | 231 | 622 | 30 | 514 | 19 |
|  | **Cotton** | 34 | 4054 | 20 | 4481 | 3 | 2830 |
| **2015** | **Corn** | 1123 | 39 | 656 | 13 | 466 | 3 |
|  | **Cotton** | 69 | 4246 | 36 | 4548 | 1 | 2796 |

Classification accuracy of cotton was above 99.81% in the same-year tests and 99.64% in cross-year tests. The accuracy of corn was as low as 91.81% in cross-year tests in 2016. The difference in classification accuracies was partly due to how corn is sown after the harvest of wheat, so a late harvest of wheat may shift the growth of corn in different years. On the other hand, the plantation of cotton is not dependent on other agricultural activities.

Confusion matrix of the Kansas dataset is depicted in Table 7. The same year user's and producer's accuracies of corn and soybean are above 98%. However, cross year accuracies are lower (Table 8). VDTW mislabel 22.84% of corn fields trained with 2018 data and tested with 2017 and 13.98% of soybean fields trained with 2018 data and tested with 2017 data. Crops in 2018 were sown eight days earlier on average compared to 2017. This caused lower accuracies in the cross-year tests.

Table 7. Average Confusion Matrix of 100 tests for VDTW Classification with 50 samples in the Kansas dataset. Columns are observations while rows are predictions. Years in rows are training and years in rows are test years.

|      |         | 2017 |         | 2018 |         |
|------|---------|------|---------|------|---------|
|      |         | Corn | Soybean | Corn | Soybean |
| 2017 | Corn    | 2048 | 142     | 2283 | 429     |
|      | Soybean | 78   | 2855    | 24   | 2630    |
| 2018 | Corn    | 1679 | 31      | 2224 | 76      |
|      | Soybean | 497  | 3016    | 33   | 2933    |

Table 8. Average User's Accuracy and Producer's Accuracy of VDTW classification results with 50 samples for the same-year and cross-year for the Kansas Dataset.

|            | User's Accuracy |         | Producer's Accuracy |         |
|------------|-----------------|---------|---------------------|---------|
|            | Corn            | Soybean | Corn                | Soybean |
| Same-year  | 98.33           | 99.04   | 98.67               | 98.79   |
| Cross-year | 91.17           | 92.48   | 88.06               | 92.47   |

Same year user's and producer's accuracies are between 98.33-99.04% (Table 8). However, the cross-year user's and producer's accuracies are up to 10% lower. Kappa values were 0.97 for the same-year and 0.81 for the cross-year tests.

**5. Discussion**

Test results show that the proposed approach improved overall accuracy results in both the same-year and cross-year tests. VDTW fuses advantages of both DTW and SAM methods; thus, it provides flexibility in time and measurement variations: DTW has the ability to be flexible in time; SAM is robust to illumination changes and measurement differences.

Previous work had an overall accuracy difference of 10% between same-year and cross-year (Zhong, Gong, and Biging 2014). The proposed approach also improved same-year crop mapping accuracies in the Harran Plain compared to previous object-based (Alganci et al. 2014) and multi-temporal (Celik, Sertel, and Ustundag 2015) studies. Our results with the Kansas dataset was also in conjunction with the previous work(Zhong, Gong, and Biging 2014) having 9% to 10% accuracy difference between the same-year and the cross-year tests. Yearly change of cropping practices decreased the accuracy of all classification methods in the Kansas dataset. On the other hand, VDTW method was more robust compared to other methods in the cross-year tests. TWDTW approach was proposed to improve DTW performance (Maus et al. 2015). However, it did not include changes in illumination and variations in measurements as in SAM or VDTW approach. Deep LSTM's accuracy was improved as the number of training samples were increased. This result was expected as DL requires a large amount of data and fine-tuning of parameters. RF with a double-sigmoid features approach has similar results compared to SAM and DTW methods.

Our multiyear crop mapping approach overcame difficulties in cross-year classification. In addition to SG data smoothing, we included interpolation of vegetation index values of cloudy data samples. This cloud information cloning approach improved cross-year overall accuracies.

NDVI and EVI were commonly used in phenological feature extraction (de Souza et al. 2015; Pan et al. 2015). However, we propose the use of MSAVI Since we obtained higher the cross-year overall accuracies with the use of the MSAVI. Soil adjusted vegetation indices, such as SAVI, include the effect of the soil line as a parameter; on the other hand, MSAVI computes the soil line parameter automatically. For this reason, the use of MSAVI further reduced variations in observation angles.

A limited time window version of VDTW, PVDTW, achieved similar overall accuracies with fewer data. PVDTW enables mid-season crop classification and has efficient computation requirements. The partial time window method may be applied to other classification algorithms such as DTW and SAM under the proposed multi-year crop mapping approach.

VDTW and PVDTW methods are not as vulnerable as the other methods to the paucity of available training data. This property is useful since an operational system can use pure phenologies (as low as a single time-series signature) or it can still operate sufficiently with fewer temporal data samples.

The proposed methods can also be extended to the classification of other crops, such as discrimination of wheat-barley, corn-soybean (Massey et al. 2017), and rice-corn (Tang et al. 2018), which have overlapped phenological phases.

The difference of the first derivative of vegetation index (VI) was evaluated as an alternative to angles between VI time-vectors. The correlation between two distinct vectors, which have different values and the same slopes, were different. VDTW incorporates Euclidean similarity implicitly, thus resulting in better discrimination.

We consider missing data acquisitions in large time windows may lower multi-year crop mapping performances. These time windows are growth and harvest where the changes are exponential rather than linear. We suggest that the curve fitting with the double logistic function or other non-linear methods may eliminate this problem.

According to our investigations in the Harran Plain, farmers may re-sow cotton if the seedlings did not emerge due to drought or heavy rains. In this case, the growth of the cotton crop was delayed, and its phenology resembled that of corn. Another issue is the growing of cotton as the second crop. However, this practice is not common and may produce low crop yields (Çopur and Yuka 2016).

One last challenge for VDTW method is that it requires more computation power than both DTW and SAM methods. Compared to the DTW, vector dot products are computed at each point instead of a simple absolute distance operation. However, VDTW achieved high performance with fewer training samples. We also suggest using the median of training samples to generate crop mapping from training data for time-sensitive or large-scale applications.

In this study, vector dynamic time warping (VDTW), an improved version of DTW, was developed and presented in a multi-year crop mapping approach for efficiently classifying crops with similar phenologies, such as corn and cotton, and other crops with distinct phenologies. The proposed method is based on optimal time vector alignment of crop phenologies for overcoming the difficulties experienced in previous efforts. Vector dynamic time warping (VDTW) for crop mapping is robust against spectral and temporal shifts in yearly crop growths.

We tested our method with multiple crops and in separate regions yielding high classification accuracies. Classification of corn and cotton, which are investigated in this study is challenging due to the overlaps in their phenological characteristics. Corn and Soybean in Kansas have partially overlapping phenologies; however phenology of crops in 2018 shifted considerably compared to 2017. The proposed VDTW method provided the highest same-year and cross-year overall classification accuracies. Our tests with the Kansas dataset showed that there is still room for improvement in cross-year crop mapping.

Another improvement of our work is employing discriminative regions for efficient crop classification PVDTW method uses optimal time window selection to achieve comparable accuracies of its base method, with less temporal data. Optimal time-periods to discriminate these crops are determined by our algorithm.

Both VDTW and PVDTW methods achieved higher classification accuracy compared to other methods with a limited number of training samples, thus reducing the repeated effort of collecting ground samples.

We believe that the proposed methods can also be expanded to classify other types of crops. Besides, the VDTW method may also be adapted to different research areas (e.g., data mining and speech recognition) where DTW is commonly preferred.

We also believe that the approach developed is highly suitable for crop mapping at regional scales. However, further additional datasets are required to expand the VDTW to countrywide levels. In the meantime, we think that the proposed approach may be used to improve the accuracy of the Ministry of Agriculture and Forestry's National Registry of Farmers in the near future for the crop types taken into consideration in this study.

## 6. Conclusions

In this study, vector dynamic time warping (VDTW), an improved version of DTW, was developed and presented in a multi-year crop mapping approach for efficiently classifying crops with similar phenologies, such as corn and cotton, and other crops with distinct phenologies. The proposed method is based on optimal time vector alignment of crop phenologies for overcoming the difficulties experienced in previous efforts. Vector dynamic time warping (VDTW) for crop mapping is robust against spectral and temporal shifts in yearly crop growths.

We tested our method with multiple crops and in separate regions yielding high classification accuracies. Classification of corn and cotton, which are investigated in this study is challenging due to the overlaps in their phenological characteristics. On the other hand, the crops in the Bismil Plain have distinct phenologies. Corn and Soybean in Kansas have partially overlapping phenologies however phenology of crops in 2018 shifted considerably compared to 2017. The proposed VDTW method provided the highest same-year and cross-year overall classification accuracies. Our tests with the Kansas dataset showed that there are still room for improvement in cross-year crop mapping.

Another improvement of our work is employing discriminative regions for efficient crop classification PVDTW method uses optimal time window selection to achieve comparable accuracies of its base method, with less temporal data. Optimal time-periods to discriminate these crops are determined by our algorithm.

Both VDTW and PVDTW methods achieved higher classification accuracy compared to other methods with a limited number of training samples, thus reducing the repeated effort of collecting ground samples.

We believe that the proposed methods can also be expanded to classify other types of crops. Besides, the VDTW method may also be adapted to different research areas (e.g., data mining and speech recognition) where DTW is commonly preferred.

We also believe that the approach developed is highly suitable for crop mapping at nationwide scales. However, further additional datasets are required to expand the VDTW to countrywide levels. In the meantime, we think that the proposed approach may be used to improve the accuracy of the Ministry of Agriculture and Forestry's

National Registry of Farmers in the near future for the crop types taken into consideration in this study.